\pdfoutput=1

\documentclass[11pt]{article}

\usepackage[]{EMNLP2023}

\usepackage{times}
\usepackage{latexsym}
\usepackage{graphicx}
\usepackage{cleveref}

\usepackage[T1]{fontenc}
\usepackage[utf8]{inputenc}

\usepackage{microtype}

\usepackage{inconsolata}

\title{nanoT5: A PyTorch Framework for Pre-training and Fine-tuning \\ T5-style Models with Limited Resources}

\author{Piotr Nawrot \\ University of Edinburgh \\ \url{https://github.com/PiotrNawrot/nanoT5}
}

\renewcommand{\footnotesize}{\fontsize{5pt}{11pt}\selectfont} 

\begin{document}
\maketitle

\begin{abstract}
State-of-the-art language models like T5 have revolutionized the NLP landscape, but their computational demands hinder a large portion of the research community. To address this challenge, we present nanoT5, a specially-optimized PyTorch framework for efficient pre-training and fine-tuning of T5 models. Drawing on insights from optimizer differences and prioritizing efficiency, nanoT5 allows a T5-Base model to be pre-trained on a single GPU in just 16 hours, without any loss in performance. With the introduction of this open-source framework, we hope to widen the accessibility to language modelling research and cater to the community's demand for more user-friendly T5 (Encoder-Decoder) implementations. We make our contributions, including configurations, codebase, pre-training insights, and pre-trained models, available to the public.
\end{abstract}

\section{Introduction}
The transformative power of large pre-trained language models such as GPT-3 \cite{Brown2020LanguageMA}, T5 \cite{Raffel2019ExploringTL}, and PaLM \cite{Chowdhery2022PaLMSL} is undeniable. However, their massive computational requirements remain a barrier for many researchers. Notably, models like T5 require extensive datasets and significant computational resources for their pre-training \cite{Raffel2019ExploringTL}. Furthermore, many open-source implementations lean heavily on TPU accelerators \cite{Shazeer2020GLUVI}, which are not as available to the academic community as GPUs.

Recognizing this gap, we introduce nanoT5, a resource-efficient, open-source PyTorch framework designed for the pre-training and fine-tuning of T5 models. Inspired by pioneering efforts such as nanoGPT \cite{karpathy2021nanogpt} and Cramming \cite{Geiping2022CrammingTA}, nanoT5 uniquely concentrates on enhancing the training pipeline specifically for T5 encoder-decoder models. Our framework includes optimized configurations and scripts, enabling researchers to pre-train a T5-Base model with 248M parameters on a single GPU in just 16 hours. Every facet, from data preprocessing and model architecture to the learning rate schedule, has been tuned for both efficiency and adaptability. With nanoT5, users can seamlessly initiate model pre-training within minutes of accessing our GitHub repository.

This paper underscores two main innovations: First, we delve into the nuances between the Adam and Adafactor optimizer performances as detailed in \cite{pretrainingT5dutch}, suggesting a version of AdamW \cite{Loshchilov2017DecoupledWD}, augmented with matrix-wise learning rate scaling based on root mean square. This variant showcases better speed and robustness compared to the default Adafactor \cite{Shazeer2018AdafactorAL}. Second, we demonstrate that T5 models trained with nanoT5, housing around 250M parameters, can achieve performance akin to the publicly-available checkpoints while requiring 150x less pre-training data.

Our primary motivation stems from the growing demand for reproducible and tuned baselines \citep{Kaddour2023NoTN}, enabling fast and small-scale hypothesis validation in the evolving realm of large pre-trained Transformers. With nanoT5, we address a gap highlighted by community requests \footnote{\texttt{https://github.com/google-research/text-to-text-transfer-transformer/issues/172}}\footnote{\texttt{https://github.com/huggingface/transformers/issues/18030}}\footnote{\texttt{https://github.com/huggingface/transformers/issues/5079}}, providing an approachable platform for working with T5 (Encoder-Decoder) architecture. To our understanding, nanoT5 pioneers the effort to reproduce T5 v1.1 pre-training using PyTorch, deviating from prior Jax/Flax implementations. We invite the community to explore our training configurations, codebase, and pre-trained models, all of which are available at \url{https://github.com/PiotrNawrot/nanoT5}.

\section{Related Work}
The landscape of open-source repositories tailored for efficient pre-training of Transformer language models is vast. Notably, nanoGPT \cite{karpathy2021nanogpt} sheds light on decoder-only models, while Cramming \cite{Geiping2022CrammingTA} homes in on the optimal pre-training of the encoder-only BERT architecture \cite{Devlin2019BERTPO}. Contrastingly, with nanoT5, we sought to bridge the existing gap by providing a standalone research template tailored for the T5-style (Encoder-Decoder) models.

To expedite the training process of nanoT5 we incorporated various optimizations. These encompass mixed precision training \citep{Micikevicius2017MixedPT}, compiled runtimes \citep{Narang2021DoTM}, and more. Additionally, we delved into the potential of efficient training methodologies such as recent optimizers \citep{Chen2023SymbolicDO, Liu2023SophiaAS}, and fast attention mechanism \citep{Dao2022FlashAttentionFA}, which are elaborated further in \Cref{sec:efficiency}. It's crucial to note that while we evaluated various efficient algorithms, we consciously opted against those, such as \citep{Nawrot2022EfficientTW, Shazeer2017OutrageouslyLN}, that would modify the core model structure. Instead, our intent with nanoT5 was to cultivate a straightforward baseline for further research endeavors. The standout contribution of our work in terms of efficient training algorithms is the AdamW variant, with the RMS matrix scaling, which improves T5 pre-training convergence.

\section{Methodology}

Our validation strategy seeks to replicate the T5-base pre-training outcomes detailed in \citep{Shazeer2020GLUVI} and the fine-tuning results of Tk-Instruct on the Super Natural-Instructions (SNI) meta-dataset \citep{Wang2022SuperNaturalInstructionsGV}.

\subsection{Training pipeline}

We've devised a comprehensive training pipeline prioritizing efficient data management, low-level optimizations, and coding simplicity, all while preserving the core model and training logic:

\begin{itemize}

\item \textbf{Dataset Handling:} Given the extensive volume of the C4 dataset, which exceeds 300GB, our repository implements concurrent data downloading with model training. This optimization speeds up the commencement of T5 model pre-training to a few minutes.

\item \textbf{Exposure and Simplicity:} Our methodology aims to strike a balance between adaptability and abstraction. With tools such as the HuggingFace Accelerator \citep{accelerate}, we abstract tasks like checkpoint management and tensor operations. Experiment tracking is realized via neptune.ai \citep{Neptune_team_neptune_ai_2019}, and we've employed hydra \citep{Yadan2019Hydra} for coordinated hyperparameter handling.

\item \textbf{Efficiency:} We've leveraged the optimizations of PyTorch 2.0 \citep{Paszke_PyTorch_An_Imperative_2019}, and employed mixed-precision training in line with established optimization guidelines \footnote{\texttt{https://huggingface.co/docs/transformers/perf\_train\_gpu\_one}}\footnote{\texttt{https://pytorch.org/tutorials/recipes/recipes/tuning\_guide.html}}.

\item \textbf{Flexibility:} Our repository is designed with adaptability in mind, offering support for multi-GPU training, gradient accumulation, and gradient checkpointing. This ensures users can reproduce our results on a variety of GPUs beyond the A100 and can experiment with configurations larger than the T5-base size emphasized in this study. Additionally, we provide support for both CPUs and Apple's ARM M1 chips.

\end{itemize}

\subsection{Pre-training}
\label{sec:pretraining-config}

Our experiments strictly follow the T5-v1.1-base training configuration \citep{Shazeer2020GLUVI}, where the model itself comprises of roughly 248M parameters. The C4 dataset \citep{Raffel2019ExploringTL}, sourced directly from Huggingface, undergoes tokenization via the Wordpiece tokenizer \citep{schuster2012japanese}, with the original model's vocabulary. During pre-processing, 15\% of input data is masked using sentinel tokens, setting the neural network's target as the prediction of these tokens, leveraging its decoder. Consistent with the original study, we've set the batch size at 128 examples, with inputs of 512 tokens and outputs of 114 tokens. Optimization is facilitated through the Adafactor optimizer \citep{Shazeer2018AdafactorAL}, combined with the Inverse-Square-Root (ISR) learning rate schedule. The model is trained for $2^{16}$ steps. For more details please refer to the original work.

\subsection{Fine-tuning}

Our fine-tuning employs the Super Natural-Instructions (SNI) meta-dataset \citep{Wang2022SuperNaturalInstructionsGV}, which has been previously used for fine-tuning models like FlanT5 \citep{Chung2022ScalingIL}, BLOOM \citep{Scao2022BLOOMA1}, and Tk-Instruct \citep{Wang2022SuperNaturalInstructionsGV}. To assess the correctness of our fine-tuning setup, and the efficiency of our pre-training, we decided to reproduce the Tk-Instruct methodology.

\begin{figure}[t]
    \centering
    \includegraphics[width=\columnwidth]{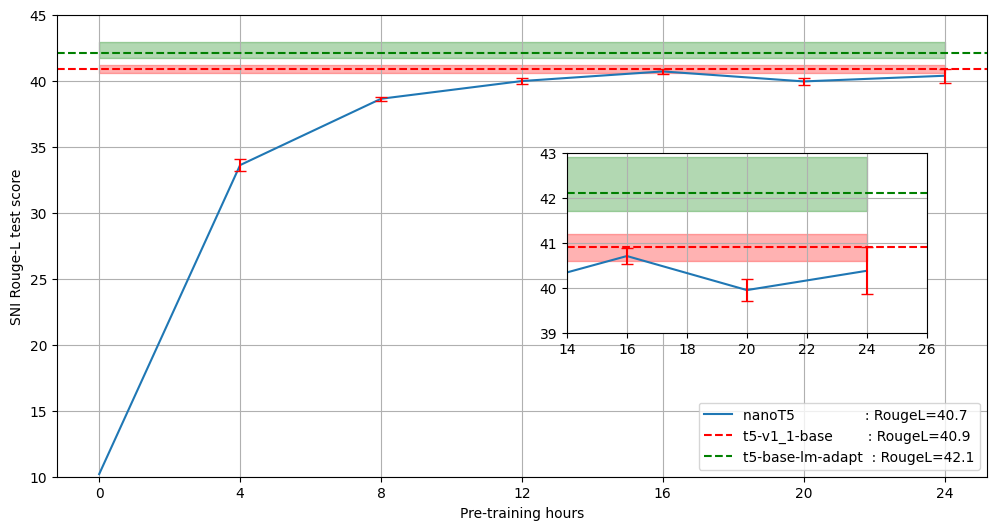}
    \caption{Downstream performance of models across various pre-training durations, including existing T5-base variants accessible through Huggingface Hub.}
    \label{fig:downstream_perf}
\end{figure}

\subsection{Reproducibility}
Ensuring that our work can be reliably replicated is a core focus of our methodology. To facilitate this, we have taken the following measures:

\begin{itemize}
\item \textbf{Model Weights:} We make the model's weights available on the HuggingFace Hub. These can be downloaded and used for fine-tuning on the SNI dataset with nanoT5.

\item \textbf{Loss Curves:} We openly share both the pre-training and fine-tuning loss curves to provide insight into the model's learning dynamics.

\item \textbf{Hyperparameters:} All hyperparameters used in our experiments have been released.

\item \textbf{Environment and Hardware:} In our repository we offer comprehensive instructions on how to recreate our environment, including detailed information about our hardware. This encompasses specifications of our CPU and GPU, as well as the relevant driver versions.

\item \textbf{Statistical Robustness:} To ensure the validity of our results, each experiment was conducted three times with different random seeds.

\end{itemize}

\section{Results}

\subsection{Reproducing Pre-Training}

By following the original experimental setup described above, we achieved a negative log-likelihood of $1.995$ on the held-out set, which is slightly inferior to the reference.

In exploring alternative optimization methods, we tested the AdamW optimizer as a potential replacement for the original Adafactor. While AdamW theoretically promises greater training stability by directly estimating the second moment of the gradients (as opposed to Adafactor's low-rank approximation), our training with AdamW diverged. This behavior mirrors findings from a study on T5 pre-training \cite{pretrainingT5dutch}. Upon further investigation, we identified that matrix-wise learning rate (LR) scaling using its root mean square (RMS) \footnote{For more details please refer to \cite{Shazeer2018AdafactorAL}, Section 8, titled "Relative Step Size"} was the crucial element ensuring Adafactor's convergence. After augmenting AdamW with this extra LR scaling, which we will refer to as RMS scaling, it not only converged but also exhibited improved stability during pre-training and operated slightly faster, thanks to the direct retrieval of the second moment from memory instead of approximating it.

\begin{figure}[t]
    \centering
    \includegraphics[width=\columnwidth]{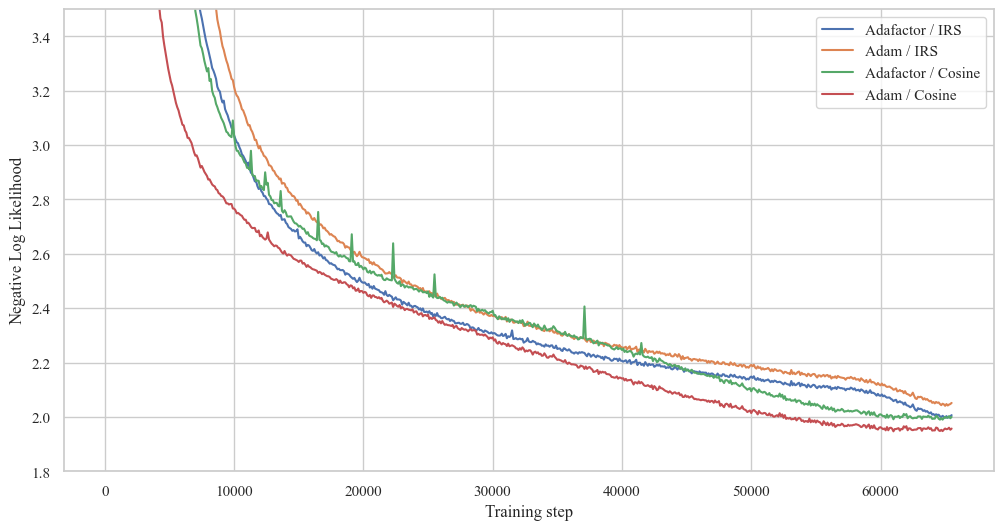}
    \caption{Training loss curves contrasting different optimizers and learning rate schedules.}
    \label{fig:pt_training}
\end{figure}

Nonetheless, when combined with the Inverse-Square-Root LR schedule, AdamW's performance was still outpaced by Adafactor. By replacing the ISR schedule with a Cosine LR Schedule, we achieved a superior negative log-likelihood of 1.953 on the held-out set, significantly surpassing Adafactor with the ISR schedule. The specific results of these experiments can be found in Table \ref{tab:pt_valid}. A comparison of the training loss curves using different optimizers (Adafactor vs. AdamW) and schedules (ISR vs. Cosine) is provided in Figure \ref{fig:pt_training}.

\begin{table*}[t]
    \centering
    \resizebox{\textwidth}{!}{
    \begin{tabular}{|c|c|c|c|c|}
        \hline
        \textbf{Mixed Precision} & \textbf{Torch 2.0 compile} & \textbf{Grad Acc} & \textbf{Time per 1 Pre-training step} & \textbf{Total Pre-training time} \\
        \hline
        FP32 & No & 2 & $\sim$4.10s & $\sim$74.6h \\
        \hline
        TF32 & No & 2 & $\sim$1.39s & $\sim$25.3h \\
        \hline
        BF16 & No & 2 & $\sim$1.30s & $\sim$23.7h \\
        \hline
        TF32 & Yes & 2 & $\sim$0.95s & $\sim$17.3h \\
        \hline
        BF16 & Yes & 1 & $\sim$0.56s  & $\sim$10.2h \\
        \hline
    \end{tabular}
    }
    \caption{Efficiency metrics across various configuration settings during pre-training, with the "Total Pre-training Time" column referencing $2^{16}$ steps following the default config.}
    \label{tab:efficient}
\end{table*}

\begin{table}[h]
    \centering
    \begin{tabular}{|c|c|c|}
    \hline
    & \textbf{Inverse-Square-Root} & \textbf{Cosine} \\
    \hline
    \textbf{Adafactor} & 1.995 & 1.993 \\
    \hline
    \textbf{AdamW} & 2.040 & \textbf{1.953} \\
    \hline
    \end{tabular}
    \caption{Comparison of negative log-likelihood scores on the held-out set of C4 using different optimization methods and learning rate schedules.}
    \label{tab:pt_valid}
\end{table}

\subsection{Fine-Tuning Performance Across Different Pre-Training Durations}

Our fine-tuning configuration strictly aligns with that of Tk-Instruct. However, there remains some ambiguity regarding whether Tk-Instruct was initialized from a regular checkpoint (google/t5-v1\_1-base) or from a version specifically tailored for Language Modelling (google/t5-base-lm-adapt). To cover all bases, we evaluated both, and successfully reproduced the original results.

Figure \ref{fig:downstream_perf} presents a performance comparison of the model we trained in various time increments (ranging from 4 to 24 hours) against the original T5-base-v1.1 model weights from Huggingface Hub and its language modeling-adapted version. Notably, our model, trained for 16 hours on a single GPU, lagged by only 0.2 RougeL on average compared to the original T5-base-v1.1. This is an impressive result given the vast disparity in training data (the T5 paper indicates training on approximately 150x more data than we did). The language modeling-adapted checkpoint outperformed both the original T5-base-v1.1 model and ours, but this language modeling model adaptation extends beyond the scope of this study. A single fine-tuning step in our setup took approximately 0.18s, culminating in roughly an hour for the entire fine-tuning process.

\subsection{Efficiency Statistics}

Table \ref{tab:efficient} showcases the efficiency metrics from our pre-training experiments. It details the time taken for a single pre-training step and the overall pre-training time based on our default configuration desribed in \Cref{sec:pretraining-config}. A noteworthy observation is that, because of the large batch size (128) used for pre-training, for numerical precisions other than BF16 we need to increase the number of gradient accumulation steps from 1 to 2.

\paragraph{Attempts at Boosting Efficiency}
\label{sec:efficiency}

In our pursuit of efficiency, we experimented with various strategies, albeit with limited success:

\begin{itemize}
    \item \textbf{Optimization Algorithms}: We assessed the performance of recent optimizers like Lion \cite{Chen2023SymbolicDO} and Sophia \cite{Liu2023SophiaAS}. However, neither outperformed the AdamW with RMS scaling.
    \item \textbf{Positional Embeddings}: We tried replacing T5's learned relative positional embeddings with ALiBi \cite{Press2021TrainST}. Although such a switch had the potential to reduce the number of parameters, leading to faster training and inference rates, and paving the way for integrating Flash Attention \cite{Dao2022FlashAttentionFA} (currently limited to non-parametric bias), our trials revealed that training with ALiBi was more volatile and yielded suboptimal pre-training loss.
    \item \textbf{FP16 Precision}: Unfortunately, all our attempts using FP16 precision consistently diverged.
\end{itemize}

\section{Conclusions}
In this study, we demonstrated the feasibility of pre-training a substantial model like T5 under resource constraints, specifically using a single A100 GPU within a 24-hour timeframe. Through selection of optimization methods and configurations, we achieved results comparable to large-scale training settings. Our intention in sharing the codebase, configurations, and training logs is to bridge the gap between research accessibility and computational resource limitations in the NLP domain. We invite and welcome community suggestions to further refine and enhance our approach.

Moving forward, we aim to enrich our codebase by incorporating additional training objectives, such as those suggested by \cite{Tworkowski2023FocusedTC, Tay2022UL2UL}, in hopes of further optimizing the training pipeline.

\section*{Acknowledgements}
This work was supported by the UKRI Centre for Doctoral Training in Natural Language Processing, funded by the UKRI (grant EP/S022481/1) and the University of Edinburgh, School of Informatics and School of Philosophy, Psychology \& Language Sciences.

\bibliography{anthology,custom}
\bibliographystyle{acl_natbib}

\end{document}